\documentclass{article}

\usepackage{graphics} 
\usepackage{epsfig} 
\usepackage{amsmath} 
\usepackage{amssymb}  
\usepackage{todonotes}
\usepackage[labelfont=bf]{caption}
\usepackage{subcaption}
\usepackage{csquotes}
\usepackage{booktabs}
\usepackage{float}
\usepackage{hyperref}
\usepackage{wrapfig}
\usepackage[margin=1in]{geometry}

\newcommand{\ie}{\emph{i.e.}, } 

\usepackage{etoolbox}

\renewcommand\footnotemark{}

\title{\Large \bf
Design and Experiments with LoCO AUV:\\ A Low Cost Open-Source Autonomous Underwater Vehicle\textbf{*}
}

\author{Chelsey Edge$^{1}$\\ \and Sadman Sakib Enan$^{1}$\\ \and Michael Fulton$^{1}$\\ \and Jungseok Hong$^{1}$\\ \and Jiawei Mo$^{1}$\\ \and \break Kimberly Barthelemy$^{2}$\\ \and Hunter Bashaw$^{3}$\\ \and Berik Kallevig$^{4}$\\ \and Corey Knutson$^{5}$\\ \and Kevin Orpen$^{2}$\\ \and Junaed Sattar$^{6}$
\thanks{Authors presented in alphabetical order of surname within academic positions due to collaborative contribution.}
\thanks{1. PhD Student/Candidate, University of Minnesota Computer Science}
\thanks{2. Undergraduate, University of Minnesota Aerospace Engineering}
\thanks{3. Undergraduate, Clarkson University Computer Science}
\thanks{4. Undergraduate, University of Minnesota Mechanical Engineering}
\thanks{5. Undergraduate, University of Minnesota - Duluth Computer Science}
\thanks{6. Assistant Professor of Computer Science at University of Minnesota}
\thanks{* Supported by the Minnesota Robotics Institute Seed Grant and the National Science Foundation award \#00074041.}
}
\date{\vspace{-7ex}}

\begin{document}

\maketitle
\pagestyle{plain}

\begin{abstract}
In this paper we present LoCO AUV, a Low-Cost, Open Autonomous Underwater Vehicle.  
LoCO is a general-purpose, single-person-deployable, vision-guided AUV, rated to a depth of 100 meters.
We discuss the open and expandable design of this underwater robot, as well as the design of a simulator in Gazebo. 
Additionally, we explore the platform's preliminary local motion control and state estimation abilities, which enable it to perform maneuvers autonomously.
In order to demonstrate its usefulness for a variety of tasks, we implement a variety of our previously presented human-robot interaction capabilities on LoCO, including gestural control, diver following, and robot communication via motion.
Finally, we discuss the practical concerns of deployment and our experiences in using this robot in pools, lakes, and the ocean.
All design details, instructions on assembly, and code will be released under a permissive, open-source license. 
\end{abstract}
\section{Introduction}

Autonomous underwater vehicles (AUVs) are a key tool in scientific and industrial work in marine and aquatic environments. 
They are used to explore shipwrecks~\cite{foley_archeologoy_2009}, chart biological habitats~\cite{williams_habitat_2010}, destroy subsea mines~\cite{sariel_mine_2006}, inspect and repair undersea architecture such as pipelines or cables~\cite{petillot_pipeline_2002}, and to do a plethora of other tasks.
However, AUVs are often expensive, large, and difficult to deploy, limiting their use to well-funded research groups, such as oceanography institutes, underwater robotics research labs, and well-funded University groups. 
Additionally, AUVs are often not sold commercially and those which are, tend to be expensive to acquire and maintain.  
There are a great number of areas where the need for AUVs is outweighed by the challenges outlined above, such as use in marine science at a state and local government level, education at lower-funded universities and secondary schools, and in hobbyist development.
If an AUV were available at a lower cost, with less overhead in deployment, and with less constraint on additions and modifications to the platform, those in these under-represented groups would be able to leverage the capabilities of an underwater autonomous agent to achieve their goals in research and education. This would also significantly reduce the barrier of entry into the theories and practices of autonomous underwater robotics.

In their sixty-some year history, autonomous underwater vehicles (AUVs) have opened an \enquote{eye into the deep} for humanity, revealing secrets of undersea life, patterns in currents, shipwrecks, and much more~\cite{williams_habitat_2010, foley_archeologoy_2009}. 
The work of oceanography institutions has propelled AUVs from their beginnings as military tools into critical scientific instruments. 
The AUVs developed by these institutions carry complex sensor payloads, and are typically either deployed in concert with a support ship or, in the case of gliders, are sent on long-term missions on their own.
While these, often large and expensive, AUVs have played an important role in the development of the field of underwater robotics, a new breed of AUV is on the horizon. 
Due to improvements in battery technology and rapid growth in consumer-marketed remotely operated vehicles (ROVs), it is now feasible to create an AUV comprised of almost entirely off-the-shelf parts, for a fraction of the cost of other AUVs.
This will open up the world of autonomous underwater robots to a new group of users who did not have the funds or capability to build or purchase other AUVs. 
Students at many levels of education will be able to use these new AUVs to learn basic principles of underwater robotics, local and state-level marine research groups will have a new ability to use AUVs in their research, and even hobbyists could start to venture into AUVs at a higher level.
We seek to contribute to this new wave of AUV by presenting an AUV well suited to the small, lower-funded teams common in student groups and local and state researchers.

\begin{wrapfigure}{r}{0.5\textwidth}
    \centering
    \includegraphics[width=0.9\linewidth]{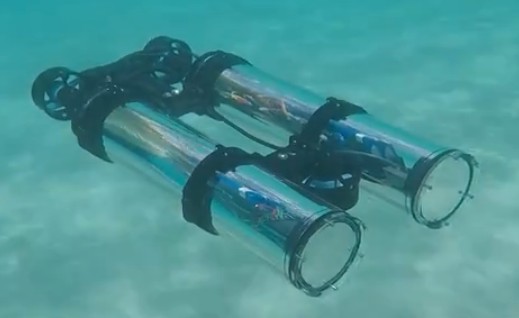}
    \caption{LoCO in the Caribbean Sea near Barbados.}
    \label{fig:loco_bbds}
    \vspace{-2mm}
\end{wrapfigure}

To this end, we present the \textit{LoCO} AUV, a \textbf{Lo}w \textbf{C}ost, \textbf{O}pen, Autonomous Underwater Vehicle. LoCO is a human-portable, easy to deploy AUV built from approximately \$4,000 worth of parts, largely off-the-shelf electronics and additively manufactured (\ie 3D printed). 
While LoCO is not as specialized as more expensive AUVs, it boasts an impressive array of capabilities at this early stage in development, from autonomous swimming in defined patterns to diver following and control by hand gestures. 
Its hardware is designed to run vision-based algorithms, often employing deep learning for perception, enabling a great number of applications in the future.
Moreover, the LoCO platform's design lends itself well to modularity and modification: the addition of new sensor payloads or other hardware would be a simple enough matter, depending on the needs of the payload. 
With the openness of the design overall, this AUV is designed to be a tool for those without access to the funds and resources required to create and deploy typical AUVs.


In this paper we make the following contributions:

\begin{itemize}
    \item We introduce the LoCO AUV: a low-cost, open AUV, designed to be easily built, modified, and deployed by small, low-resource teams.
    \item We discuss the current status of LoCO's system design, including its electrical, computing, and sensor systems.
    \item We discuss the current status of LoCO's state estimation and local control algorithms.
    \item We present a Gazebo-based simulator for LoCO, for prototyping algorithms before deployment.
    \item We port our research lab's existing human-robot interaction capabilities to LoCO from the Aqua AUV.
    \item We will release all code, plans, 3D printed part models, assembly instructions, and documentation under a permissive, open source license at \url{https://loco-auv.github.io/}.
\end{itemize}
\section{Related Work}
We present a brief summary of previous work which is related to LoCO.
There are numerous AUVs and papers dealing with AUVs in order to provide an exhaustive review. Therefore, the authors primarily highlight works dealing with small and low-cost AUVs.
AUV development can be considered to have begun~\cite{gafurov_autonomous_2015} in 1957 with the development of the Special Purpose Underwater Research Vehicle (SPURV), at the University of Washington, funded by the United States Office of Naval Research~\cite{widditsch_spurv_1973}.  SPURV was used until 1979, and followed by SPURV II~\cite{nodland_spurv_1981}, which improved on the hydrodynamic design of its predecessor, and increased the number of sensors on-board. 
Soon after, the REMUS AUV~\cite{allen_remus_1997} was developed by Woods Hole Oceanographic Institute, in an attempt to develop a smaller, lower-cost AUV. 
The efforts put forward for REMUS are, in many ways, a spiritual predecessor to LoCO, as the team developing REMUS focused on low cost and a small support staff.
Later, REMUS 600~\cite{stokey_development_2005} improved on the design of REMUS, with a side-looking synthetic aperture sonar and improved endurance and payload flexibility. 
By this time, gliders such as Seaglider~\cite{eriksen_seaglider_2001} and Deepglider~\cite{osse_deepglider_2007} were setting the standard for a new type of AUV: the long endurance glider, with mission times measuring in weeks and months rather than hours.
While traditional torpedo-shaped AUVs and gliders still dominated, innovation in the realm of mini-AUVs (mass of 20-100kg) and micro-AUVs (mass of 20kg or less) began to spring up, with the flippered Aqua~\cite{dudek_aqua_2007} being an example of the variation now existent in the field.

In the last five years, development in the field of micro-AUVs has been particularly interesting, with the development of HippoCampus~\cite{hackbarth_hippocampus_2015} and SEMBIO~\cite{amory_sembio_2016}, micro-AUVs for swarm applications, as well as AUVs with less typical drive designs, such as a momentum-drive single actuated robot, which rotates its inner body to move the exterior passive flaps and produce swimming motion~\cite{goncalves_design_2016}.
Other AUVs such as SHAD~\cite{goncalves_design_2016}, HOBALIN~\cite{okamoto_development_2016}, and Sparus II~\cite{carreras_sparus_2018} focused on hovering motion for seabed inspection and observation tasks. 
Lastly, the development of general-purpose micro-AUVs is still going strong, with Bluefin Sandshark~\cite{underwood_design_2017}, a docking AUV~\cite{wu_test_2018}, and other similar AUVs appearing in the last few years.
The world of AUV development research remains a thriving one, full of new innovation and ideas. 
LoCO AUV attempts to lower the barrier of entry into that world even further than it has been lowered thus far, with an entirely open and reproduceable design that will be available under a permissive, open source license.
LoCO can be purchased in component parts and built, and then modified to ones own standards making it an excellent platform for groups at any level, regardless of level of experience or funding.

\begin{figure*}
    \vspace{2mm}
    \centering
    \begin{subfigure}{.24\textwidth}
        \centering
        \includegraphics[width=\linewidth, trim= 0 1cm 0 0, clip]{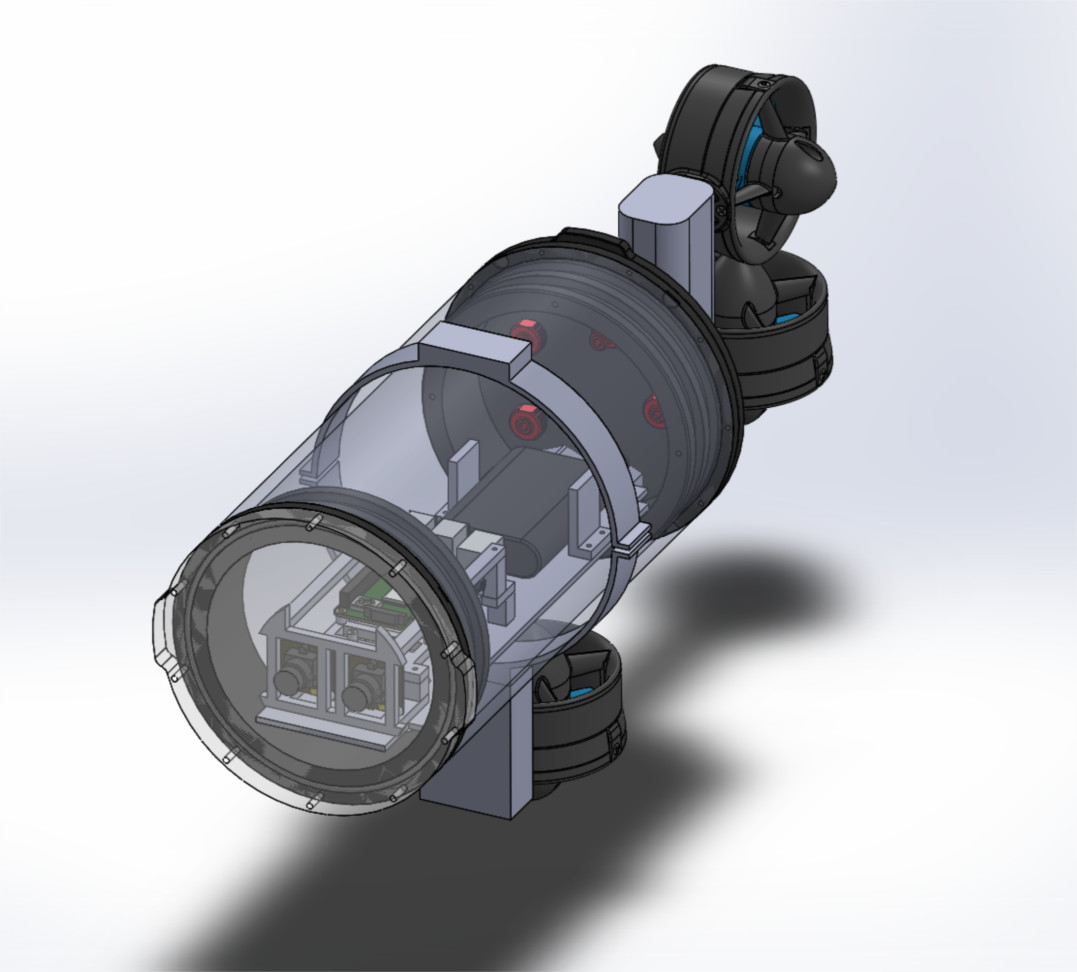}
        \label{fig:loco_developments_heli}
        \vspace{-1mm}
        \caption{\enquote{Helicopter} design.}
    \end{subfigure}
        \begin{subfigure}{.24\textwidth}
        \centering
        \includegraphics[width=\linewidth, trim= 0 0.5cm 0 0, clip]{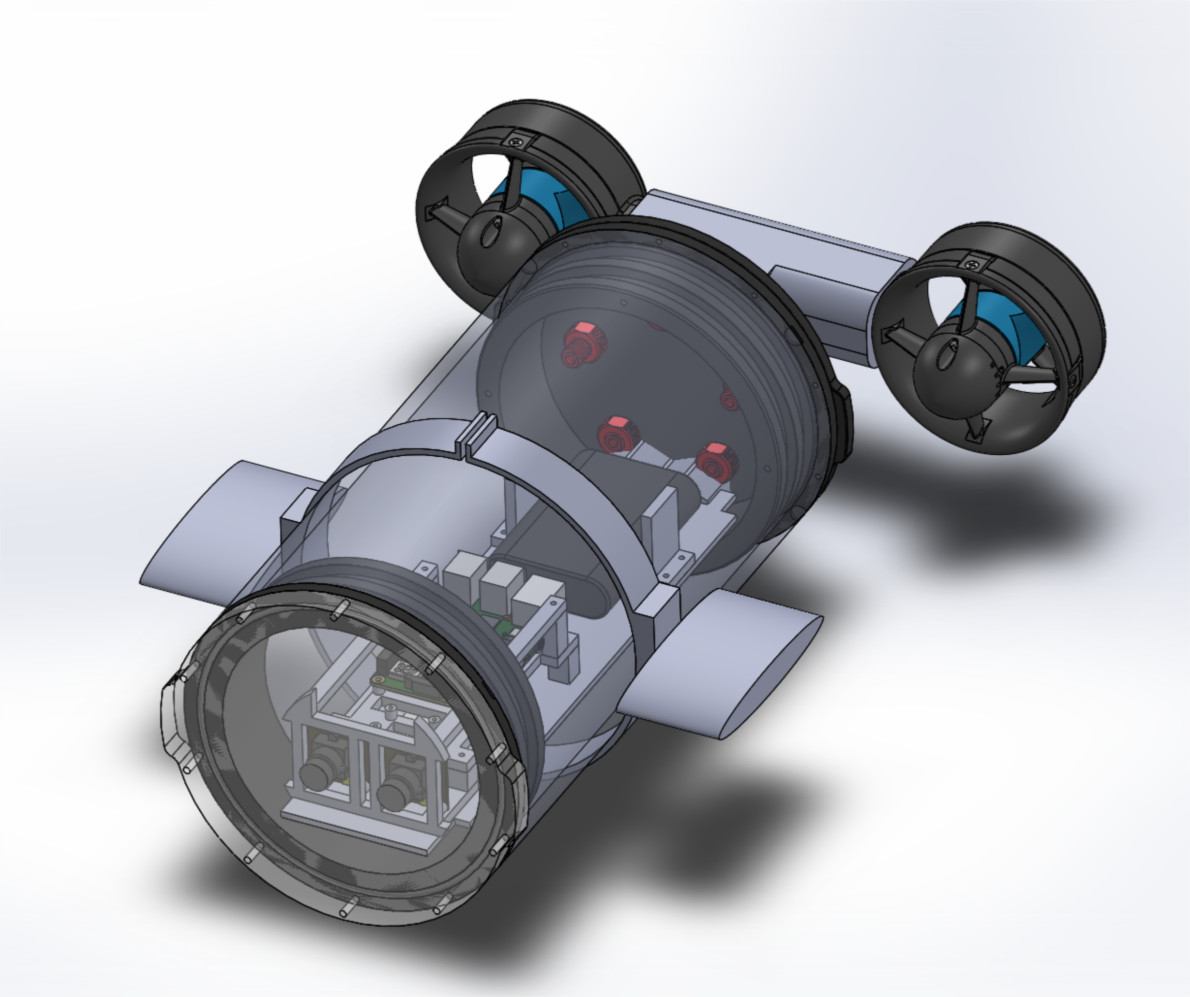}
        \label{fig:loco_developments_sub}
        
        \caption{\enquote{Submarine} style.}
    \end{subfigure}
        \begin{subfigure}{.24\textwidth}
        \centering
        \includegraphics[width=\linewidth, trim= 0 0 0 0.0cm, clip]{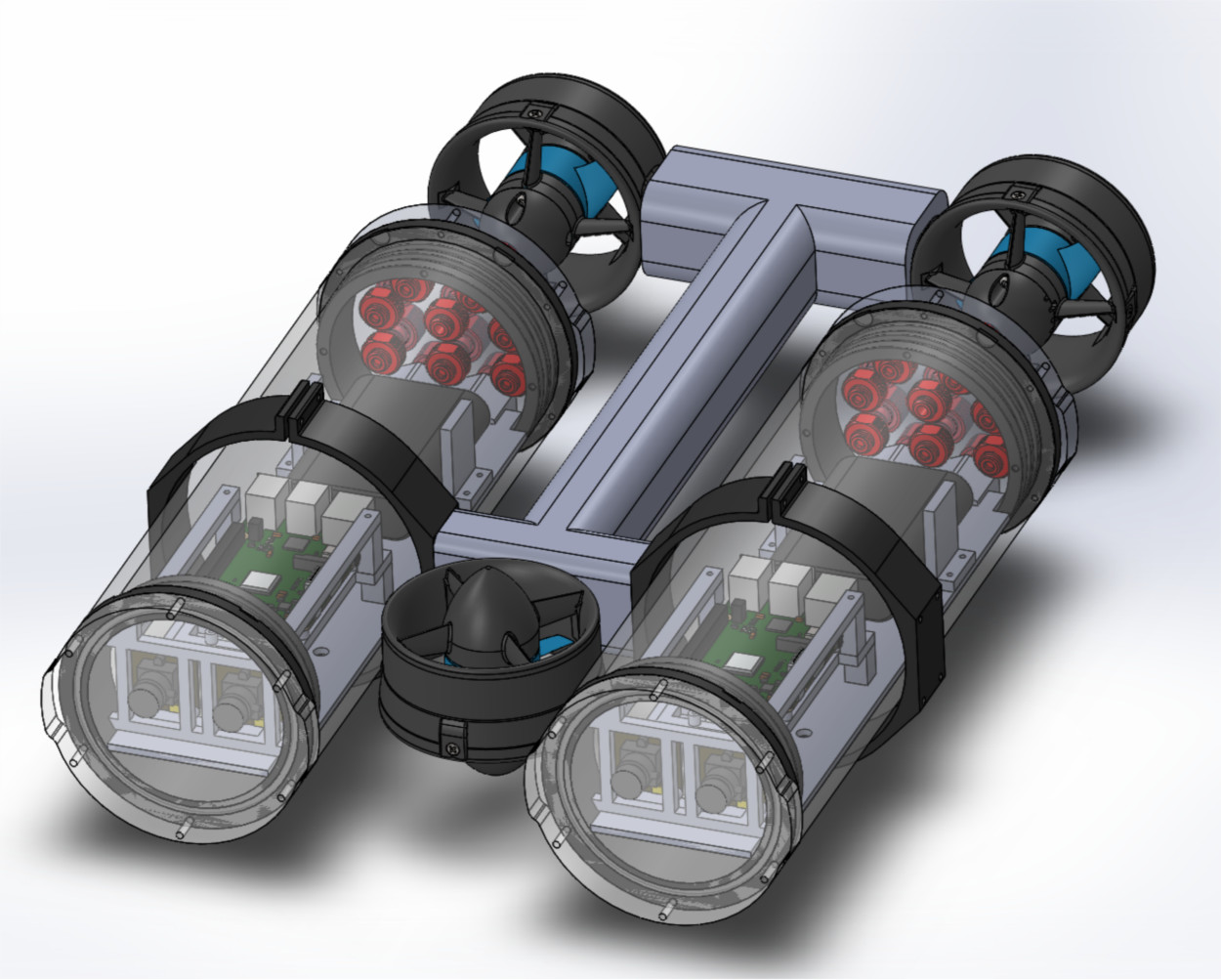}
        \label{fig:loco_developments_two_tube}
        \vspace{0mm}
        \caption{\enquote{Binocular} design.}
    \end{subfigure}
        \begin{subfigure}{.24\textwidth}
        \centering
        \includegraphics[width=\linewidth, trim= 0 -4cm 0 0, clip]{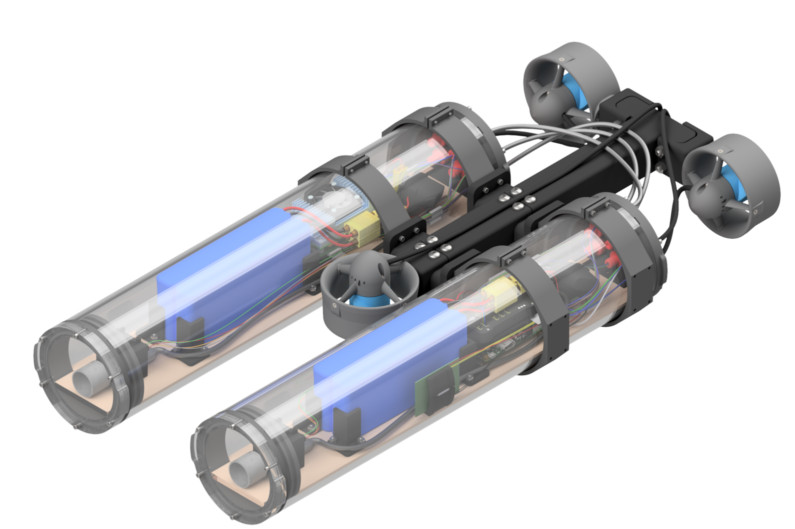}
        \label{fig:loco_developments_final}
        \vspace{-2mm}
        \caption{Final design.}
    \end{subfigure}
    
    \caption{CAD renderings showing the development of LoCO AUV.}
    \label{fig:loco_development}
    \vspace{-5mm}
\end{figure*}
\section{System Design}
\label{sec:system}

LoCO is designed to be an all-purpose AUV, adaptable to a variety of missions. 
The standard configuration is a dual-camera, vision guided AUV with three thrusters. 
The following subsections describe the robot's electrical, thruster control, computational, and sensor systems.

\begin{wrapfigure}[9]{r}{0.6\textwidth}
    \centering
    \vspace{-8mm}
    \begin{tabular}{ll}
        \textbf{LoCO AUV Specifications} & \\
        \toprule
         Dimensions (L x W x H) & 73.1cm x 34.4cm x 14.1 cm\\
         Weight & 12.47 kg (27 lb) \\
         Maximum Speed & 1.5 m/s \\ 
         Battery Life (Idle) & 18 hrs, 30 mins\\
         Battery Life (Average) &  2 hrs, 20 mins\\
         Battery Life (Max Thrust) & 30 mins\\
    \end{tabular}
    \caption{LoCO AUV Specifications.}
    \label{fig:loco_specs}
    \vspace{-2mm}
\end{wrapfigure}

\subsection{Overall System Layout}
\label{sec:system:overall}
\begin{figure*}
    \centering
    \begin{subfigure}{0.8\textwidth}
        \centering
        \includegraphics[width=\linewidth, trim=0 2.5cm 0 2.5cm, clip]{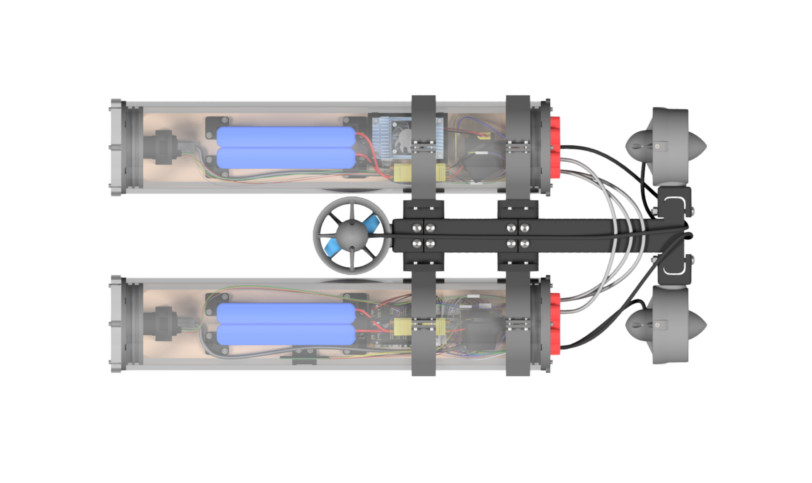}    
        \caption{}
        \label{fig:loco_design_3d}
    \end{subfigure}
    \vspace{2mm}
    \begin{subfigure}{0.8\textwidth}
        \centering
        \includegraphics[width=\linewidth, trim=0 1cm 0 0, clip]{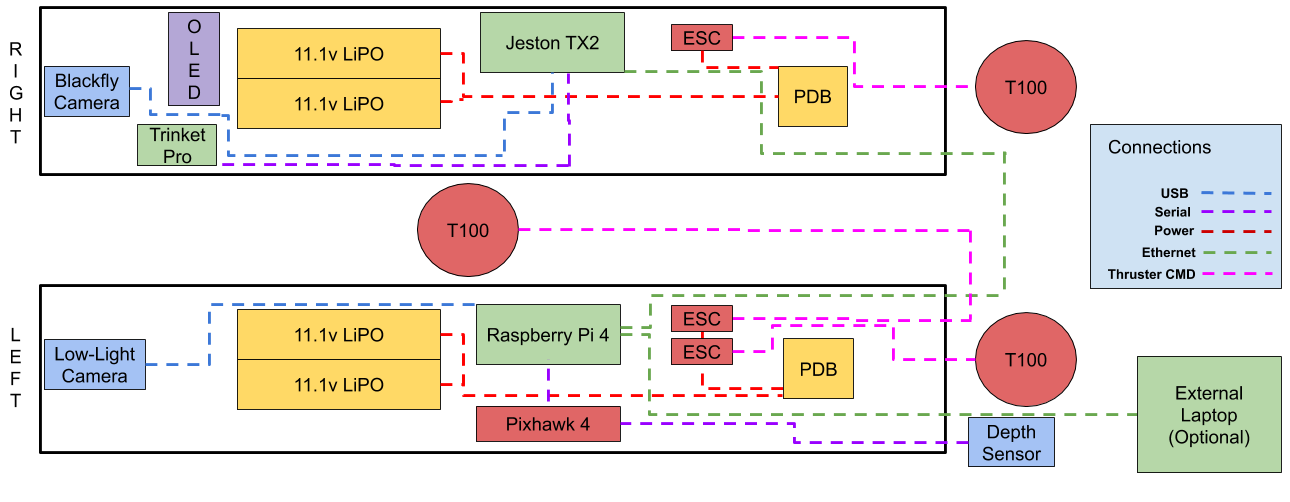}
        \caption{}
        \label{fig:loco_design_schematic}
    \end{subfigure}
    \caption{(a) Top-down view of the 3D model and (b) system schematic for LoCO AUV.}
    \label{fig:loco_design}
    \vspace{-4mm}
\end{figure*}

LoCO is comprised of two water-tight enclosures, each containing various components, with one thruster mounted between the enclosures, and two mounted behind, as seen in Figure \ref{fig:loco_design}.
While a number of designs were considered (see Fig. \ref{fig:loco_development}), the two-enclosure design was selected for a variety of reasons. 
Firstly, it allowed for a reasonable placement of a pitch-control thruster, along with providing space in between the enclosures which would be an excellent place to mount new sensors, thrusters, or manipulators in the future.
Additionally, the design called for two enclosures side by side, narrower than the enclosures used for the other designs, which would reduce the robot's forward profile, allowing it to move through the water with less resistance.
Finally, the separation into two enclosures enforces a base level of modularity.
Most control-related electronics are in the left-hand enclosure (shown in Fig. \ref{fig:loco_design_3d} and Fig. \ref{fig:loco_design_schematic}), with the computational hardware for deep learning inference in the right-hand enclosure. 
While any type of hardware modification requires changes throughout the system, changes or replacements can be made with minimal impact on the layout of components internally. 

\subsection{Watertight Enclosures and 3D printed substructure}
\label{sec:system:enclosures}
The enclosures used are cast acrylic structures with a 10.16 cm internal diameter, produced by Blue Robotics. 
The enclosures are rated to 100m depth of operation, sufficient for our needs.  
Their length is variable to any modifications that are made, but the length for our configuration is 50.8cm.
The front of each enclosure is a flat piece of cast acrylic to reduce visual distortion and increase surface area for mounting cameras, while the back cap is aluminum with openings for cables, plugged by nut-and-bolt assemblies termed penetrators.
They are linked with a custom 3D printed structure connected to aluminum clamps manufactured by Blue Robotics.
This structure is also used to mount the thrusters.

Internally, components are mounted on a piece of laser-cut medium-density fiberboard (MDF), with 3D printed mounting substructures.
The MDF is held in place within the enclosures via a 3D printed part that attaches to the penetrators poking through the back plate.
A strip of velcro beneath the MDF is used to attach 28 gram blocks of ballast cut from stainless steel bar stock.
There is also enough space under the MDF to fit bags of desiccant, which help to eliminate condensation in the event that the enclosure is sealed in humid air (common in pool and tropical environments).
This internal design is beneficial to the modularity of LoCO, as it allows for additional parts to be mounted by allocating space on the MDF and then simply attaching them with screws. 
Additionally, the external structure design is flexible, allowing for the movement of the clamps along the enclosures to fix the thrusters wherever is appropriate. 
The \enquote{backbone} 3D-printed part between the clamps can be omitted with no apparent loss of structural integrity.
It would, however, provide useful mounting space for external sensors or actuators, such as a sonar altimeter or a gripper.
\vspace{-2mm}

\subsection{Electrical Systems}
\label{sec:system:electrical}

\begin{wrapfigure}[]{r}{0.6\textwidth}
    \centering
    \vspace{-8mm}
    \includegraphics[width=\linewidth]{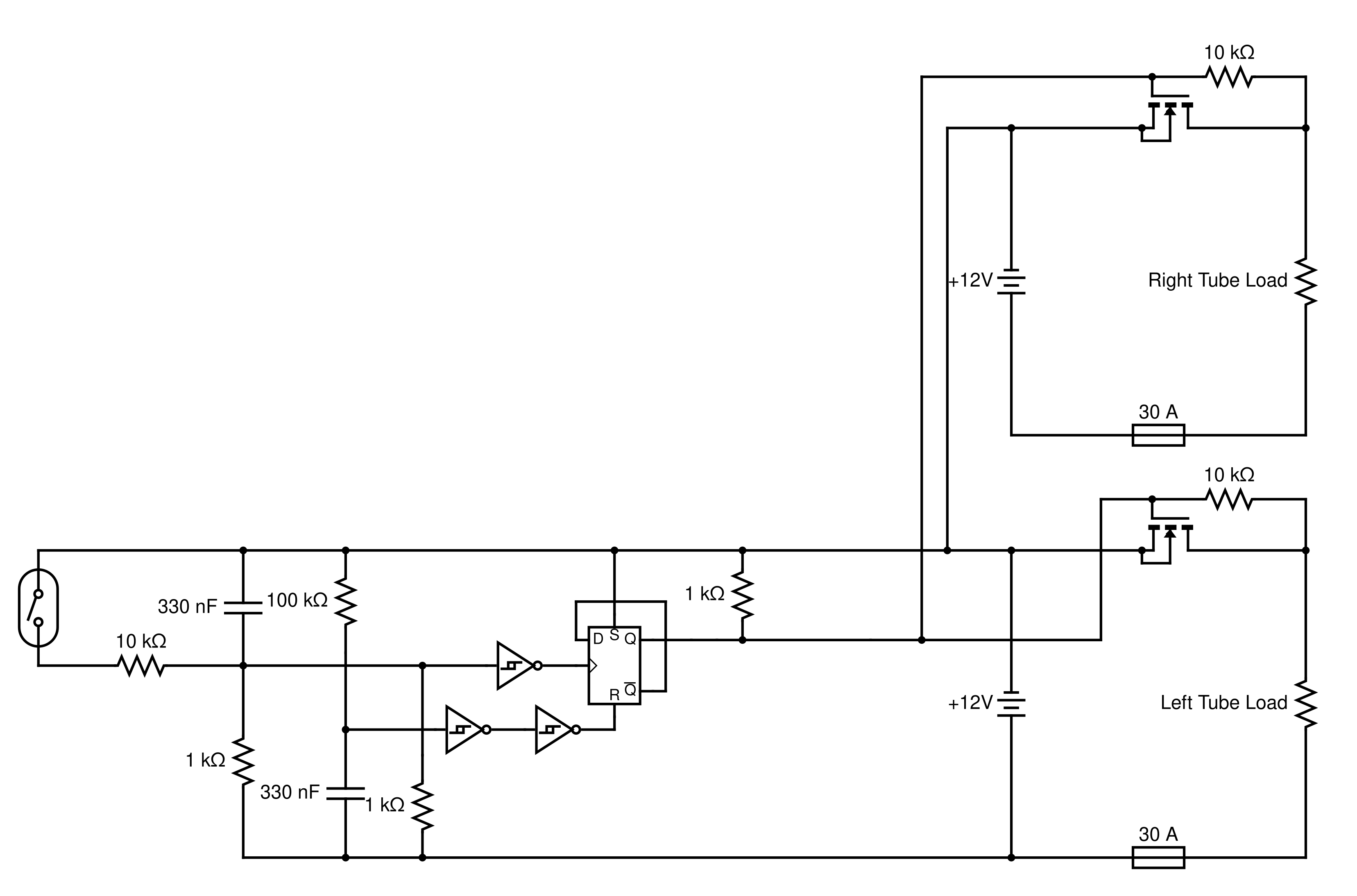}
    \caption{Power Systems Diagram.}
    \label{fig:loco_switch}
    \vspace{-2mm}
\end{wrapfigure}
LoCO's electronics are powered by four 11.1 volt lithium-polymer 8000 mAh batteries, two in each enclosure. 
The batteries in each tube are connected in parallel, providing a supply of 9.6-12.6 volts depending on their charge. 
Each tube has a low-voltage alarm which sounds when the two batteries fall below the minimum threshold of 9.6 volts.
However, a custom circuit which will measure the batteries' level of charge and report the data to LoCO's computing systems is in development.
The electrical systems in each enclosure are designed in such a way that they only power components in their respective tube.  Each pair of batteries is connected to a power distribution board (PDB) which provides six 12 volt outputs, and one 5 volt output.  
In the case of the left-hand enclosure, an external 5 volt step down converter was added in order to power the additional computing systems in that enclosure.
Despite the electrical systems of each tube being separate, a single power switch circuit controls the On/Off state of the entire robot. The circuit is situated in the left enclosure and utilizes a magnetic reed switch for input. This allows LoCO to be turned on and off using a magnetic \enquote{key}.
By using a magnetic key rather than a physical switch, we reduce possible points of failure due to leakage.  The left portion of Fig. \ref{fig:loco_switch} contains the reed switch and components that create the toggle signal, while the right side diagrams how the toggle signal is used to control the power sources.

\subsection{Thruster Control Systems}
\label{sec:system:thruster}
The thrusters employed for LoCO are Blue Robotics T100s, which use a brushless DC motor and a plastic propeller. 
Blue Robotics is discontinuing the T100, but they also produce a more powerful thruster (the T200) in the same dimensions, so future builds of LoCO AUVs will use T200s. 
These thrusters are controlled via pulse-width-modulation (PWM), which is managed by electronic speed controllers (ESCs).  
In turn, the ESCs are controlled by a Pixhawk autopilot board, employing the ArduPilot/ArduSub control software.
While we do not currently make use of all the features of the Pixhawk and ArduSub, the open software and hardware of the PX4~\cite{meier_px4_2015} and ArduPilot~\cite{ardupilot} projects lend themselves well to our goals of creating an open platform which others can contribute to and adapt to their needs.

\subsection{Computing Systems}
\label{sec:system:computing}
LoCO has two primary computer systems: a Jetson TX2 for deep learning inference and a Raspberry Pi 4 (4GB) for control.
The Jetson TX2 is mounted on a Connect Tech Orbitty carrier board for interfacing, and is largely responsible for managing processes which involve deep neural network inference. 
Due to its location in the right-hand enclosure, however, the TX2 is also responsible for managing the robot's OLED display via the connected microcontroller and processing images from the camera mounted in the right enclosure. 
In the left enclosure, the Raspberry Pi is used to process images from the camera in that enclosure and used as the controller interfacing with the Pixhawk autopilot over a serial connection.
The Jetson and the Raspberry Pi are connected via Ethernet with a Cat5e cable for a maximum throughput of 100Mbit/s.


\subsection{Vision Systems And Other Sensors}
\label{sec:system:sensors}
LoCO was designed with a stereo vision system in mind. 
The authors are currently investigating a number of cameras in order to select the appropriate camera for use in the final release version of the robot.
Currently, an FLIR Blackfly\texttrademark{} S USB3 camera is mounted in the right enclosure: a high powered camera which costs nearly \$600.  
The left enclosure currently contains a significantly less expensive USB camera from Blue Robotics, which is nearly a sixth the price. 
While the FLIR camera is more fully featured and complex, it is also a greater burden on the computational resources of the computer it is attached to, while the simpler USB camera is much more efficient.
The cameras are currently being compared for their quality, maximum feasible frame-rate, and effectiveness as part of the overall system before a decision is made. 
In addition to its vision system, LoCO employs a pressure sensor to measure its depth under the surface, which is mounted on the back plate of the left enclosure.
There is also an inertial measurement unit (IMU) contained within the Pixhawk autopilot unit. 
Lastly, while it is not a part of the design, the authors are currently working on integrating a sonar altimeter, as it is likely to be a commonly used sensor.

\section{Software}
LoCO has a variety of computing devices: a Raspberry Pi 4, an Nvidia Jetson TX2, a Pixhawk autopilot unit, and an Adafruit Trinket Pro microcontroller. 
The Raspberry Pi and Jetson TX2 both run versions of Ubuntu, a popular open source operating system, while the Pixhawk runs a real-time open source OS with ArduSub, and the Trinket is flashed directly with open-source software. 
The software which allows LoCO to function as an untethered autonomous vehicle rather than an ROV is distributed across the computing devices, and consists of a mixture of ROS packages, Ardusub, and Arduino code. 
All of this software is under some form of permissive, open source license, which makes LoCO's software stack free for users to explore, expand, and enhance.
In this section we describe a portion of LoCO's software, omitting a full description for brevity.

\subsection{Camera Drivers}
As mentioned previously in Section \ref{sec:system:sensors}, two cameras are currently being tested for use with LoCO.  
The first, a FLIR Blackfly S model, is controlled through a package from Neufield Robotics~\cite{neufield_github} which implements a Spinnaker SDK driver in ROS, the Robot Operating System~\cite{quigley_ros_2009}.
This package allows for relatively full-featured control of the camera's many parameters, as well as controlling the camera's frame capture either through software triggering, or by waiting for a signal from the camera's GPIO pins.
The other camera being tested conforms to the UVC standard, so the \verb!libuvc_camera! ROS package is used to control it. 

\subsection{State Estimation}
\begin{wrapfigure}[]{l}{0.5\textwidth}
    \centering
    \includegraphics[width=0.9\linewidth]{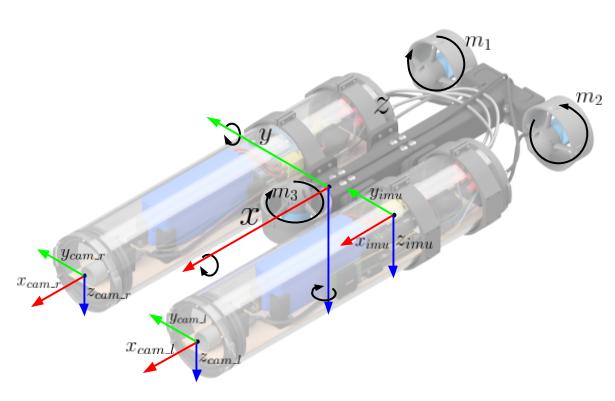}    \caption{A free-body diagram of LoCO AUV with IMU, camera, thruster, and robot frames.}
    \label{fig:loco_freebody}
    \vspace{0mm}
\end{wrapfigure}
A number of options are under development for state estimation in LoCO. 
Firstly, the Pixhawk provides an IMU-based state estimation using the Extended Kalman Filter~\cite{julier_extension_1997}.  
While this is useful, LoCO currently uses the \verb!robot_localization!~\cite{robot_localization} ROS package to estimate its orientation via IMU data directly from the Pixhawk's IMU, as this provides greater control over the tuning of the extended or unscented Kalman filters provided by the package.
Additionally, the package allows the fusion of multiple sources of information into one estimate, so if additional IMUs or sources of information became available, they could easily be integrated. Two possible sources of this information are currently in development: a downward-facing camera for monocular visual odometry and an odometry estimate based on a combination of thruster inputs and a hydrodynamic motion model of the robot.

\subsection{LoCO Pilot Controller}
In order to facilitate motion control of LoCO from a variety of sources (a teleoperation mode, or autonomous behavior algorithms), a motion control package entitled \verb!loco_pilot! has been developed. 
The package provides an interface which abstracts the control into a simple message type (\verb!\loco_pilot\Command!), containing thrust, pitch, and yaw values between $-1.0$ and $1.0$.  
This allows users to avoid the \verb!mav_ros!, MAVLink, and Ardusub systems which are required for controlling the robot, and simply publish these messages to the correct topic. 
In addition, the \verb!loco_pilot! package implements a set of motion primitives and advertises them as ROS services, allowing one to simply call a service to turn the robot to an angle, move forward or back, or follow a circle or square trajectory.
The package is under active development, adding more features and capabilities as the state estimation of the robot improves.

\subsection{Menu Control System}
Discussed in more detail in Section \ref{sec:hri:menu}, LoCO has a menu system, implemented in the ROS package \verb!loco_menu! which allows an operator to control the robot in untethered mode, by inputting commands via hand gestures or ARTags~\cite{fiala_artag_2005}. 
This menu system allows a user to select an option, which can assigned to be a variety of subroutines, including ROS service calls, ROS launch files, etc.
It uses a yaml format to define menus, making it a simple matter to create new menus for any task or environment.
\section{Simulation}
The Gazebo-based simulator~\cite{koenig_gazebo_2004} currently under development for LoCO has the potential to reduce time and money spent testing software and dynamic behavior of the robot. The Gazebo simulation utilizes ROS for robot modeling and control, allowing for an interface with current LoCO software.

\subsection{Modeling and Visualization}
\begin{wrapfigure}[]{r}{0.5\textwidth}
   \vspace{-6mm}
   \centering
   \includegraphics[width=\linewidth, trim=4cm 2cm 10cm 1cm, clip]{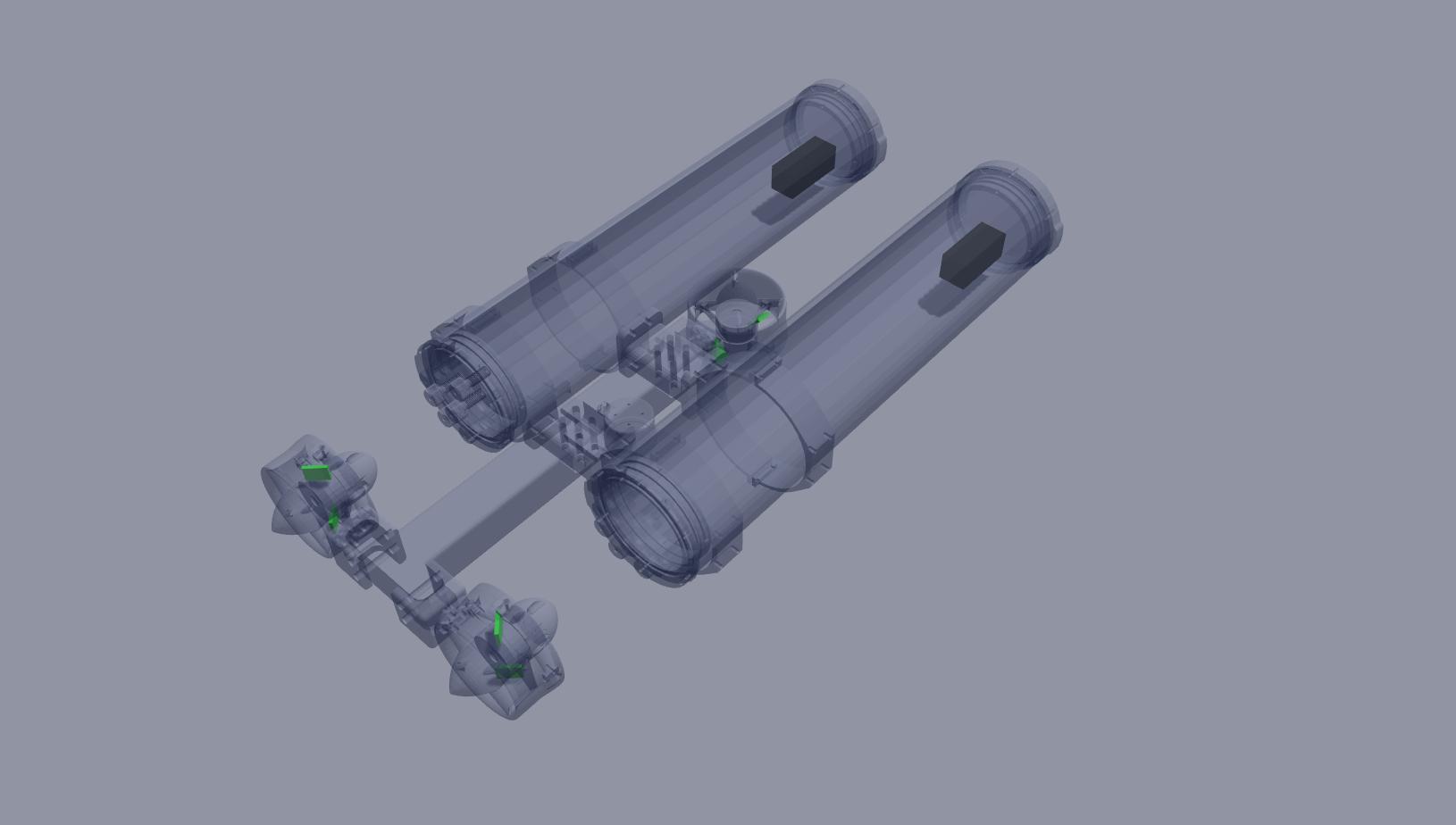}
    \caption{The LoCO AUV in Gazebo simulation.}
    \label{fig:loco_gazebo}
    \vspace{-2mm}
\end{wrapfigure}
The SolidWorks CAD design model shown in Figure \ref{fig:loco_development}d was used as the template for creating the robot's Universal Robotic Description Format (URDF) model file. LoCO’s inertial matrix and center of mass location were approximated based on a uniform density for all components using the SolidWorks Mass Analysis tool, taking into account the pre-existing mass measurements of LoCO. The center of mass was approximated between the LoCO tubes laterally and vertically, just behind the vertical thruster (see Figure \ref{fig:loco_freebody}). To improve simulation efficiency while not affecting simulation physics, the mesh file that provides the visual representation of the robot does not include internal components. The initial components modeled in the simulation are representations of the front cameras to provide future creation of sensor data, and thruster links to allow for modeling of LoCO motion. Mass and inertial properties for the cameras and thruster propellers are assumed to be negligible.

Collision properties for each link of the robot were also specified so the simulation can model physical impacts between the robot and its surroundings. 
The collision boundaries were modeled to be the same as the visual boundaries for all links except the main body of LoCO. Due to the size and complexity of the mesh, the collision boundary for the main body was modeled as a box in order to decrease simulation computational requirements.
The model is shown in Figure \ref{fig:loco_gazebo}, where the removal of internal components from the mesh can be seen.

 \subsection{Physics}
 In Gazebo, real-time fluid mechanics are not directly simulated, but rather the hydrodynamic forces are calculated, then applied to the robot. First, the ODE physics engine automatically applies a gravitational force to LoCO. Since LoCO is ballasted to be neutrally buoyant, the Gazebo BuoyancyPlugin applies the appropriate buoyancy force. The volume of LoCO is set in its model to overwrite the errors that would be produced by the bounding box approximation implemented by Gazebo to calculate model volume.
 
 There are various options available for simulating movement or propeller propulsion in Gazebo, 
 but it was determined the most accurate control method for the simulator would be directly applying thruster forces with the GazeboRosForce plugin. 
 The underwater thrust force data for the Blue Robotics T100 (and the T200) thrusters is readily available on their website, which provides for simple calculation and application of forces in Gazebo. The robot thruster links are modeled as sets of blades in case future integration of propeller dynamics is desired.
 
Determining the drag forces applied to an underwater model is a complex task. Though methods have been developed to model underwater locomotion with modifications to a typical scalar mass and inertia matrix through a Kirchhoff tensor~\cite{weissman_rigid_2012}, this would require modification to the default Gazebo physics engine. To balance the complexity of the simulation program, an approximation of underwater forces served as the preferable option. Experimental underwater trials were performed with LoCO, with the objective for the robot to reach horizontal, steady-state forward velocity in a calm environment at various thruster power inputs. IMU data was used to calculate velocity, and, with thruster forces known from the available Blue Robotics data, the balancing water drag force could be calculated, and therefore the cumulative coefficient of drag. This is used in the simulation to apply drag force against LoCO's forward and reverse motion. Since similar underwater data cannot be readily gathered for vertical, lateral, and rotational motion of the robot, those drag parameters currently remain to be evaluated. 
 
 \subsection{Control}
A simulation control node was written to bridge existing LoCO software with the Gazebo simulation. Receiving typical LoCO motion commands, it applies thrust and drag forces to the robot by using ROS Wrench messages. It also subscribes to the link states published by Gazebo so that frame transformations can be applied to desired thrust within the node before publishing the Wrench commands.

\section{Human-Robot Interface}
\label{sec:hri}
In order to demonstrate LoCO's capabilities to be used for a variety of AUV research topics, as well as to improve the usability of the platform, we ported a number of HRI capabilities previously developed for the Aqua AUV.
These capabilities require the use of LoCO's deep learning, vision guidance, and local control capabilities. 
Additionally, a strong interactive interface will be key to making LoCO usable by scientists with little programming experience.

\begin{wrapfigure}[12]{r}{0.5\textwidth}
    \vspace{-4mm}
    \centering
    \includegraphics[width=\linewidth, trim= 13cm 19cm 7cm 8cm, clip]{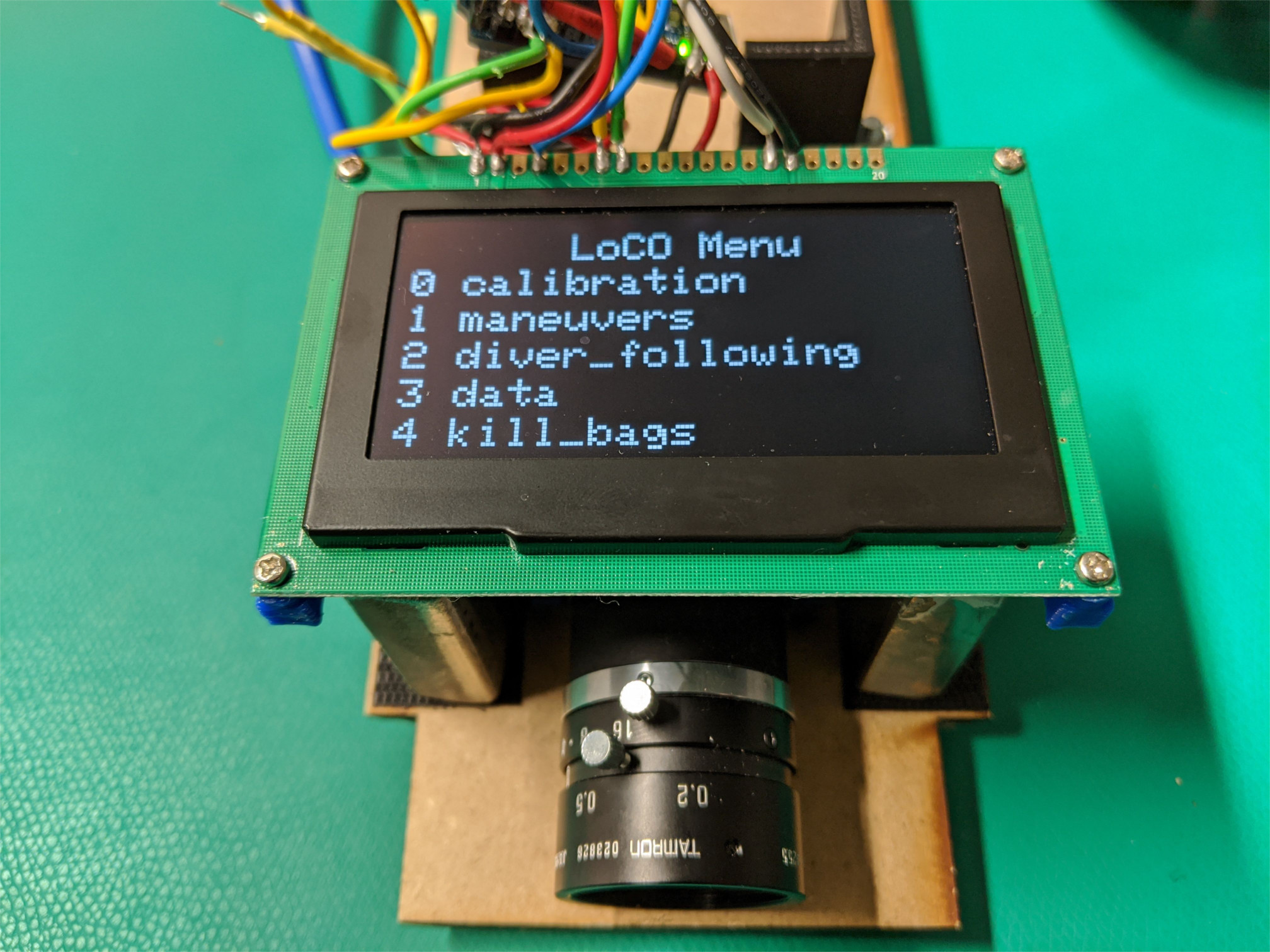}
    \caption{LoCO's OLED display showing a menu.}
    \label{fig:loco_menu}
    \vspace{0mm}
\end{wrapfigure}
\subsection{Menu System}
\label{sec:hri:menu}
LoCO's current interface is based on simple menu selection system. 
The robot's user selects from a set of 5 options using one of the methods described in the following subsection.
While these options are displayed on LoCO's OLED display for ease of use, the menu system operates independently of its display component.
These options could be connected to a variety of ROS endpoints, including running a launch file, launching a node, calling a service, terminating a node or setting a parameter. 
Menu options can also be set to submenus, allowing nesting and categorization of options.
Once a menu item has been selected, while response depends on what endpoint the option has been linked to, typically an action is taken relatively quickly, then the menu is available again once the action has been completed. 
In some cases, the option has a timeout associated with it, or must be manually canceled. 
The system is designed to be as adaptable as possible, with menu definitions being loaded in the form of yaml configuration files, making the process of writing new menu configurations as painless as possible.

\begin{wrapfigure}[]{l}{0.5\textwidth}
    \vspace{0mm}
    \centering
    \includegraphics[width=\linewidth, trim= 0 2cm 0 2cm, clip]{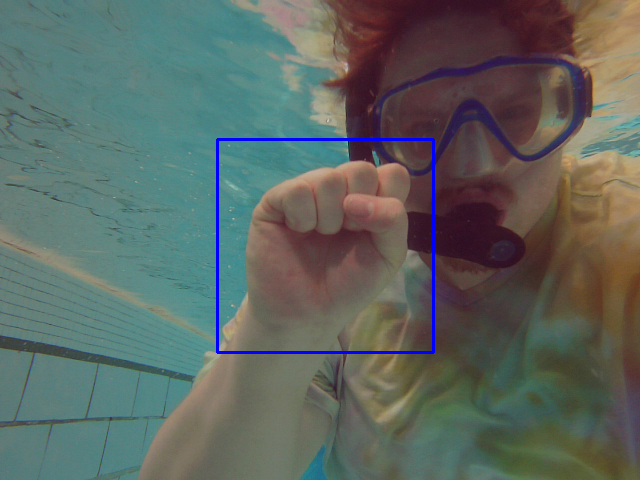}
    \caption{Gesture detection for a \enquote{0} gesture.}
    \label{fig:loco_ar+hand}
    \vspace{-0mm}
\end{wrapfigure}
\subsection{ARTags and Hand Gestures}
\label{sec:hri:input}
LoCO employs two vision-based methods to select menu items.  
The first uses small AR tags as \enquote{flashcards}. 
A ROS package based on AR Toolkit detects tags in the view of the left enclosure's camera. 
To select a menu item, one simply has to display an AR tag corresponding to the number of the item.
Alternatively, LoCO has been outfitted with a gesture recognition system, which can be used similarly. 
This gesture recognition system was initially presented for the Aqua AUV~\cite{islam_reconfiguration_2018}, and is based on a hand pose classification deep neural network.
To select a menu item, one must simply initiate a selection with the \enquote{Ok} gesture, then show the camera the number gesture for the appropriate item. 

\subsection{Robot Communication Via Motion}
\label{sec:hri:rcvm}
While the menu is displayed on the vehicle's OLED, LoCO also employs a method for robot-to-human communication previously developed for the Aqua AUV: RCVM~\cite{fulton_closing_2019}.
RCVM (robot communication via motion) is a method for robot feedback using motion as the communication vector. 
For instance, to confirm some information to its human interactant, the robot can pitch up and down, mimicking a head nodding gesture. 
RCVM enables communication at greater distances and from more accessible angles than the OLED, though it does sacrifice communication \enquote{bandwidth}.
It also serves as a good test of the local motion control systems previously described, as fast and accurate motion control is key to producing proper motion for communication.

\begin{wrapfigure}[]{r}{0.5\textwidth}
    \vspace{-6mm}
    \centering
    \includegraphics[width=\linewidth, trim= 1cm 0 3cm 5cm, clip]{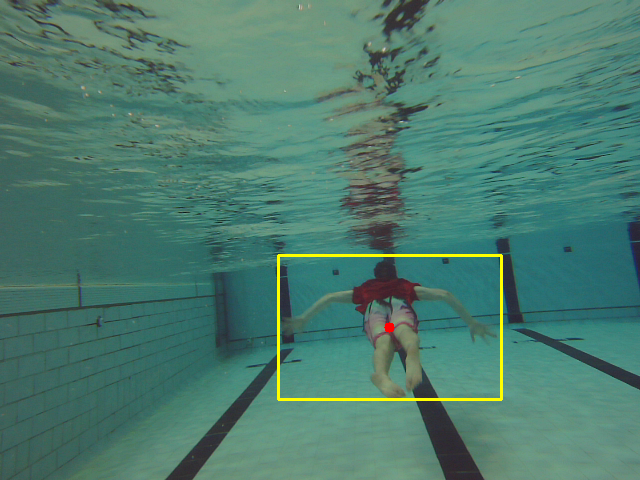}
    \caption{LoCO-eye view of following a diver.}
    \label{fig:loco_diver}
    \vspace{-2mm}
\end{wrapfigure}
\subsection{Diver Following}
\label{sec:hri:diver_following}
The final HRI capability reproduced on LoCO was diver following. 
While diver following has been explored by a variety of authors, the ability of an AUV which works with human partners to follow those partners is foundational.
Diver following can be used to convoy with a robot to a chosen location, to guide a robot through a specific route during data collection, or simply as the end goal, to name a few uses.
As with the gesture recognition system, diver following on LoCO employs a deep neural network for diver detection previously used for the Aqua AUV~\cite{islam_toward_2019}, making it an excellent use case test for the platform.
Once the detection has been made, a PID-based controller is used to generate thruster control inputs which will maneuver the bounding box into the center of the frame.
The algorithm uses bounding box size as a rough, stereo-free estimate of distance to the diver.
While there is room for improvement in the diver following in practice, the existing algorithm does a serviceable job of following.
\section{Experiment}
One of LoCO's many strengths is its ability to be deployed by a small team. 
Particularly when deploying the robot without a tether, it is easy to assemble, seal, and carry by hand to the water's edge and deploy.
In this section we will briefly discuss some of the deployment process, as well as our experiences from a variety of deployment scenarios. 

\subsection{Deployment Methodologies}
When deploying LoCO, a number of things must be done. 
Firstly, the batteries must be placed into their seats on the MDF and connected with the electrical system.
This is a simple enough affair, merely requiring the use of a few XT90 connectors. 
After this, the watertight enclosures must be closed, a straightforward process of pushing the MDF substructure into the tubes until the gaskets are seated. 
Following this, it is recommended to pump air out of the enclosures to a vacuum level of -15mm hg to ensure that there are no leaks in any of the enclosures' seal.
With a weak vacuum established, the specialized penetrators used for attaching the vacuum pump are sealed with vent plugs. 
Once that has been done, the only thing remaining is to assemble the superstructure by simply inserting four bolts through 3D printed parts and tightening their respective nuts.
With this complete, the robot is ready to be deployed.

\subsection{Deployment Scenarios Thus Far}

LoCO AUV has been deployed in pools seven times, at Wayzata Bay, Minnesota once, and off the coast of Barbados five times. 
LoCO's deployments have, thus far, been without incident. 
Of those deployments, the majority have been tethered for data collection and live debugging purposes.
There have been a handful of tetherless deployments, all in the Caribbean Sea.
Most of these deployments were undertaken with a team of 4 or fewer people.
In most cases, one person acted as \enquote{robot wrangler}, interacting with the robot and staying near it in the case of field deployments. 
The other team members were typically spread between taking external videos of the deployment, observing and debugging systems via tether, or working as the target of some algorithm (i.e. diver following).
This shows the ease with which this robot can be deployed.  
As LoCO's local motion controller and HRI suite improve, the feasibility of single-person deployments will grow exponentially. 
The primary roadblocks to a single-person deployment at the moment are the lack of diver-relative station keeping and the fact that most deployments of LoCO at the moment focus on developing some aspect of the robot, not achieving some other disconnected goal. 
The authors are in collaboration with marine biologists from the University of Minnesota to begin trials with LoCO acting as data acquisition support to their typical observation methods. 
These trials will undoubtedly provide useful insight into ways that the robot's interfaces and capabilities can be used by scientists with little-to-no robotics knowledge.

\section{Conclusions}
\begin{figure}[h]
    \vspace{2mm}
    \centering
    \begin{subfigure}{0.48\textwidth}
        \centering
        \includegraphics[width=\linewidth]{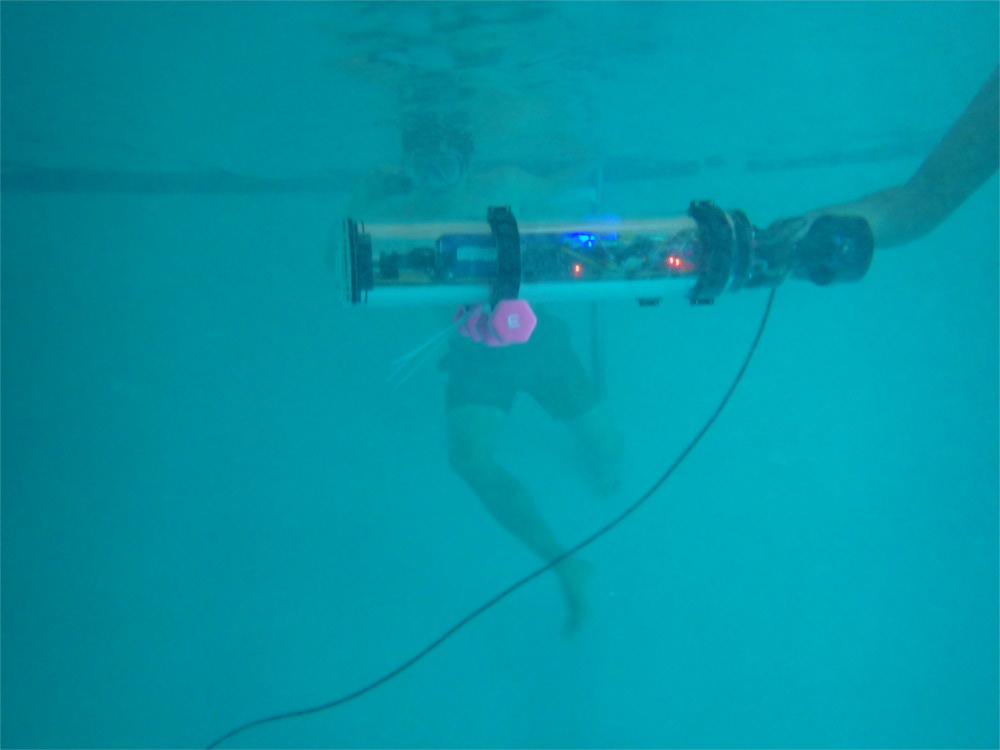}
    \end{subfigure}
    \begin{subfigure}{0.48\textwidth}
        \centering
        \includegraphics[width=\linewidth]{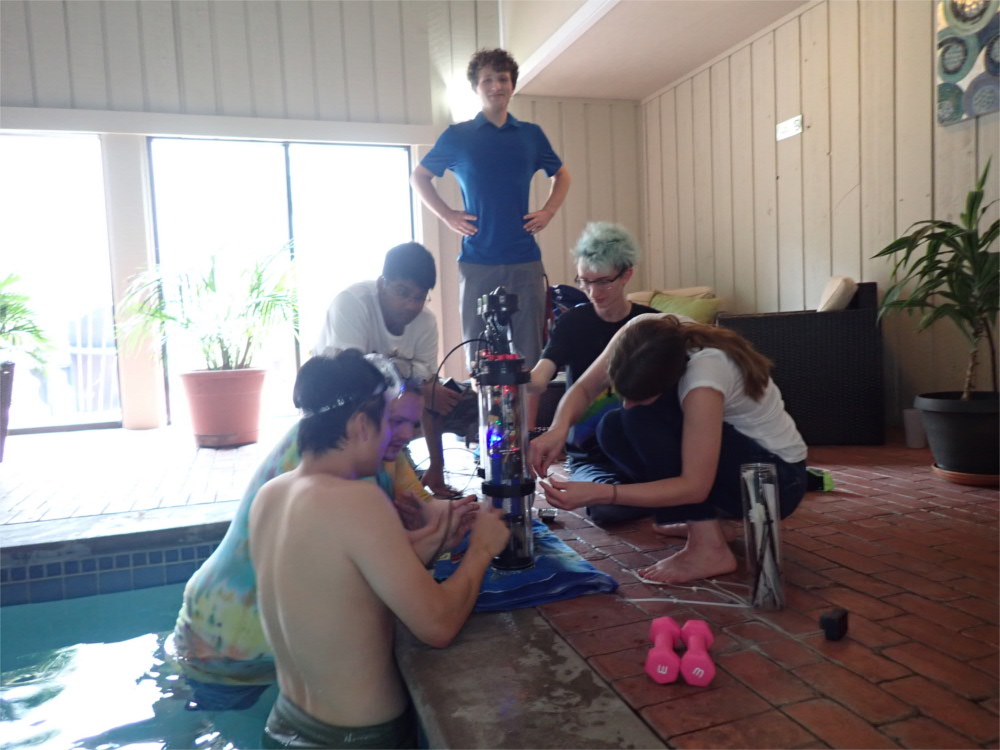}
    \end{subfigure}
    \vfill\vspace{1.5mm}
    \begin{subfigure}{0.48\textwidth}
        \centering
        \includegraphics[width=\linewidth]{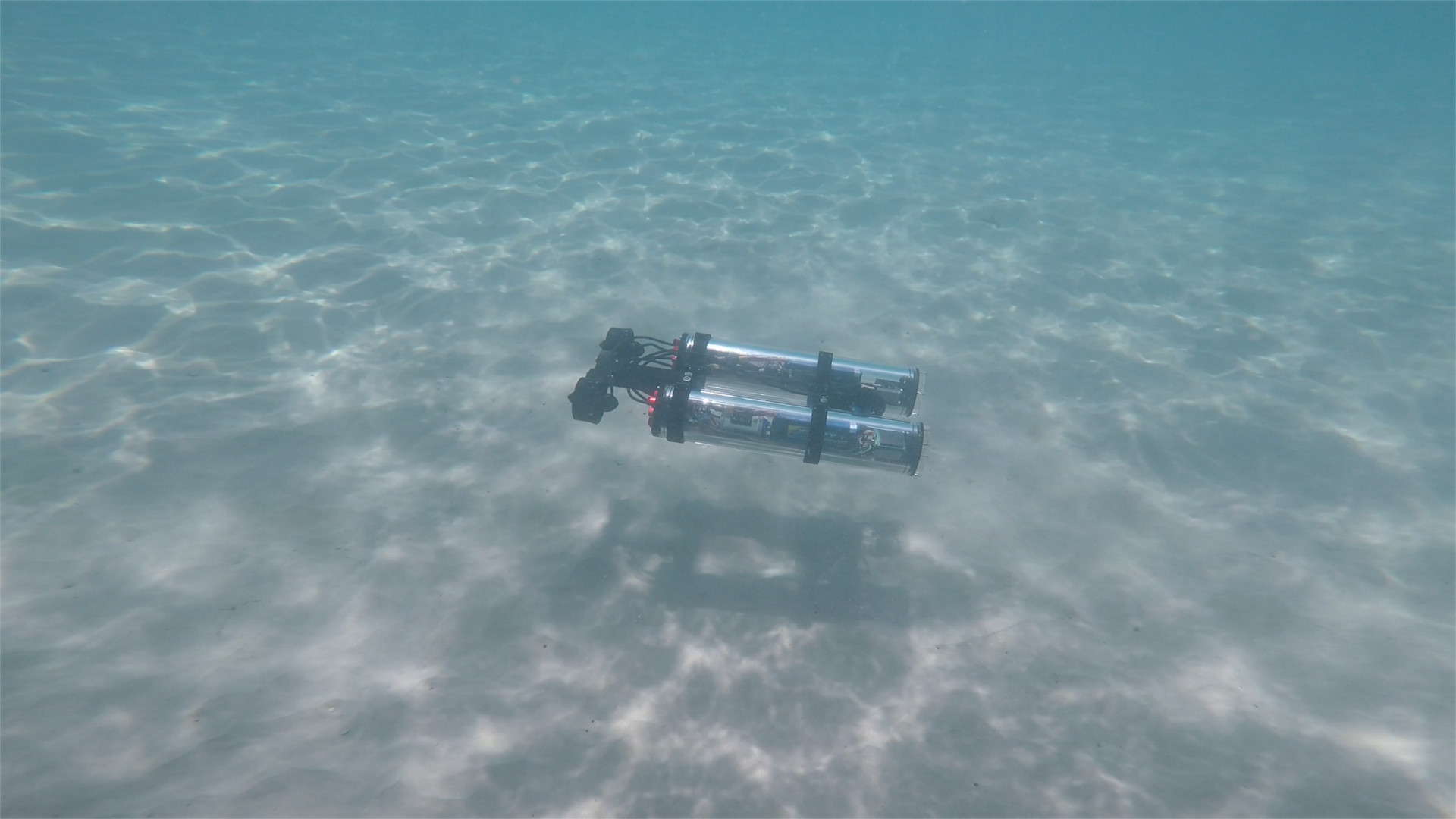}
    \end{subfigure}
        \begin{subfigure}{0.48\textwidth}
        \centering
        \includegraphics[width=\linewidth]{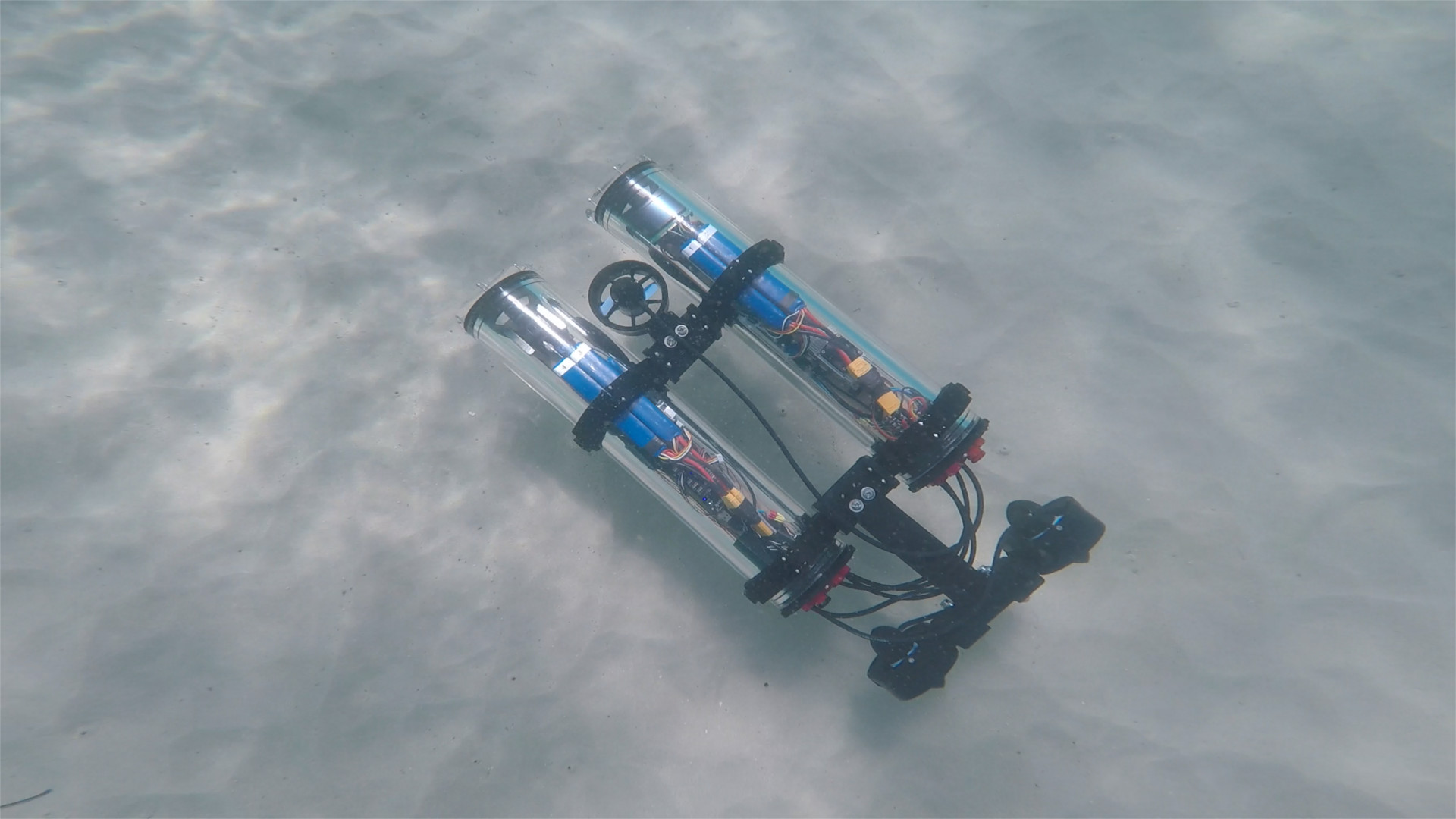}
    \end{subfigure}
    \caption{A sampling of early LoCO deployments.}
    \label{fig:loco_deployments}
    \vspace{-6mm}
\end{figure}
In this paper, we presented the design of LoCO AUV, a Low-Cost, Open Autonomous Underwater Vehicle. 
Along with details of the vehicle's hardware and software design, we discussed our development of a Gazebo-based simulator as well as an HRI interface for LoCO, and our experiences in deployment thus far.
More detailed documentation of hardware and software, as well as a parts list and assembly instructions will soon be available at \url{https://loco-auv.github.io/} under a permissive open-source license.
While no individual scientific leap forward is presented in this paper, LoCO represents the summation of many smaller pieces of progress. 
It is not the first AUV of its size or capabilities, nor is it faster or more depth capable than some AUVs. 
However, LoCO is, to our knowledge, one of the first open AUVs to be made available to the public, significantly lowering the barrier to entry in the field of robotics.
LoCO's designed purpose is and always has been to provide a platform for underwater robotics to those who could not have previously afforded their own.  
It is in this capacity that LoCO shines, and will perform with distinction. 
We hope to see LoCO robots and new variants of the platform begin to appear throughout the world, as LoCO enables more and more interested parties to take their first steps into the field of underwater robotics.

\bibliographystyle{abbrv}
\bibliography{main.bib}

\end{document}